\documentclass{svproc}
\usepackage{url}
\usepackage{helvet}         
\usepackage{courier}        
\usepackage{color}
\usepackage{makeidx}         
\usepackage{graphicx}        
\usepackage[misc]{ifsym}                            
\usepackage{multicol}        
\usepackage[bottom]{footmisc}
\usepackage{tabularx}
\usepackage{multirow}
\usepackage{natbib}
\usepackage[hidelinks]{hyperref}
\usepackage{algorithm}
\usepackage{algorithmic}
\usepackage[center]{caption}
\usepackage{longtable}
\usepackage{rotating}
 \usepackage{amsmath}
\usepackage{subfigure}
\usepackage[utf8]{inputenc}
\usepackage[T2A]{fontenc}
\usepackage[russian,english]{babel}

\hypersetup{colorlinks=true, linkcolor=green, citecolor=red, urlcolor=blue}

\begin{document}
\mainmatter              
\title{Kolmogorov Arnold Networks and Multi-Layer Perceptrons: A Paradigm Shift in Neural Modelling}
\titlerunning{Kolmogorov Arnold Networks and Multi-Layer Perceptrons: A Paradigm Shift in Neural Modelling}  
%
\author{Aradhya Gaonkar\inst{1} \and Nihal Jain\inst{2}
Vignesh Chougule\inst{3} \and Nikhil Deshpande\inst{4} \and Sneha Varur\inst{5} \and Channabasappa Muttal\inst{6}}
\authorrunning{Aradhya Gaonkar et al.} 
%
\tocauthor{Aradhya Gaonkar, Nihal Jain, Vignesh Chougule, Nikhil Deshpande, Sneha Varur, Channabasappa Muttal}
\institute{School of Computer Science and Engineering, KLE Technological University, Hubballi, 580031, India,\\
\email{01fe22bci019@kletech.ac.in, 01fe22bci022@kletech.ac.in, 01fe22bci007@kletech.ac.in, 01fe22bci017@kletech.ac.in, sneha.varur@kletech.ac.in, channabasappa.muttal@kletech.ac.in}}

\maketitle              

\begin{abstract}
The research undertakes a comprehensive comparative analysis of Kolmogorov-Arnold Networks (KAN) and Multi-Layer Perceptrons (MLP), highlighting their effectiveness in solving essential computational challenges like nonlinear function approximation, time-series prediction, and multivariate classification. Rooted in Kolmogorov's representation theorem, KANs utilize adaptive spline-based activation functions and grid-based structures, providing a transformative approach compared to traditional neural network frameworks. Utilizing a variety of datasets spanning mathematical function estimation (quadratic and cubic) to practical uses like predicting daily temperatures and categorizing wines,the proposed research thoroughly assesses model performance via accuracy measures like Mean Squared Error (MSE) and computational expense assessed through Floating Point Operations (FLOPs). The results indicate that KANs reliably exceed MLPs in every benchmark, attaining higher predictive accuracy with significantly reduced computational costs. Such an outcome highlights their ability to maintain a balance between computational efficiency and accuracy, rendering them especially beneficial in resource-limited and real-time operational environments. By elucidating the architectural and functional distinctions between KANs and MLPs, the paper provides a systematic framework for selecting the most suitable neural architectures for specific tasks. Furthermore, the proposed study highlights the transformative capabilities of KANs in progressing intelligent systems, influencing their use in situations that require both interpretability and computational efficiency.
\end{abstract}

\section{Introduction}\label{sec:intro}
Neural networks, inspired by biological systems \cite{biologicalsystems}, have transformed computing, excelling in image processing \cite{imageprocessing}, natural language comprehension \cite{nlp}, and data analysis. While mimicking human cognition has expanded machine intelligence, designing efficient architectures remains a challenge, especially in resource-limited environments. Fundamental computations like squares and cubes are crucial in scientific computing, engineering, and finance \cite{financial}. Expanding to multidimensional classification underscores efficiency needs in healthcare, logistics \cite{logistics}, and finance, driving research into computationally efficient neural architectures.

MLPs and KANs have emerged as promising solutions. MLPs, discussed by Zhang et al. \cite{MLP}, identify patterns using layered neurons across applications. KANs, introduced by Liu et al. \cite{liu2024kankolmogorovarnoldnetworks}, apply Kolmogorov’s theorem for efficient nonlinear function estimation. Their architectural advancements position them as an MLP alternative, with Janiesch et al. \cite{mldl} emphasizing their ability to streamline function approximation.

This study makes four key contributions: it compares KAN and MLP in function approximation, time-series forecasting, and classification; evaluates computational efficiency via FLOPs alongside accuracy metrics such as MSE and classification accuracy; establishes a framework for model selection based on experimental findings; and assesses KAN and MLP performance in estimating mathematical functions, handling complex datasets like the Wine dataset, and recognizing temporal trends.

Evaluation occurs across two dimensions: first, accuracy using MSE for regression \cite{mse, regression} and classification accuracy for categorical tasks; second, computational efficiency via FLOPs \cite{flops}, providing insights into resource requirements. This dual analysis bridges theory with practical applications, aiding model selection in resource-constrained environments.

The paper is structured as follows: Section 2 explores the theoretical foundations of KAN and MLP, focusing on the Kolmogorov-Arnold theorem. Section 3 outlines the methodology, detailing dataset preparation, model architecture, and evaluation metrics. Section 4 presents a comparative analysis of accuracy and computational efficiency. Finally, Section 5 summarizes findings and suggests future research directions.

\section{Background Study}\label{sec:abc}
\begin{figure*}[h]
    \centering
    \includegraphics[width=0.95\linewidth,height=0.55\linewidth]{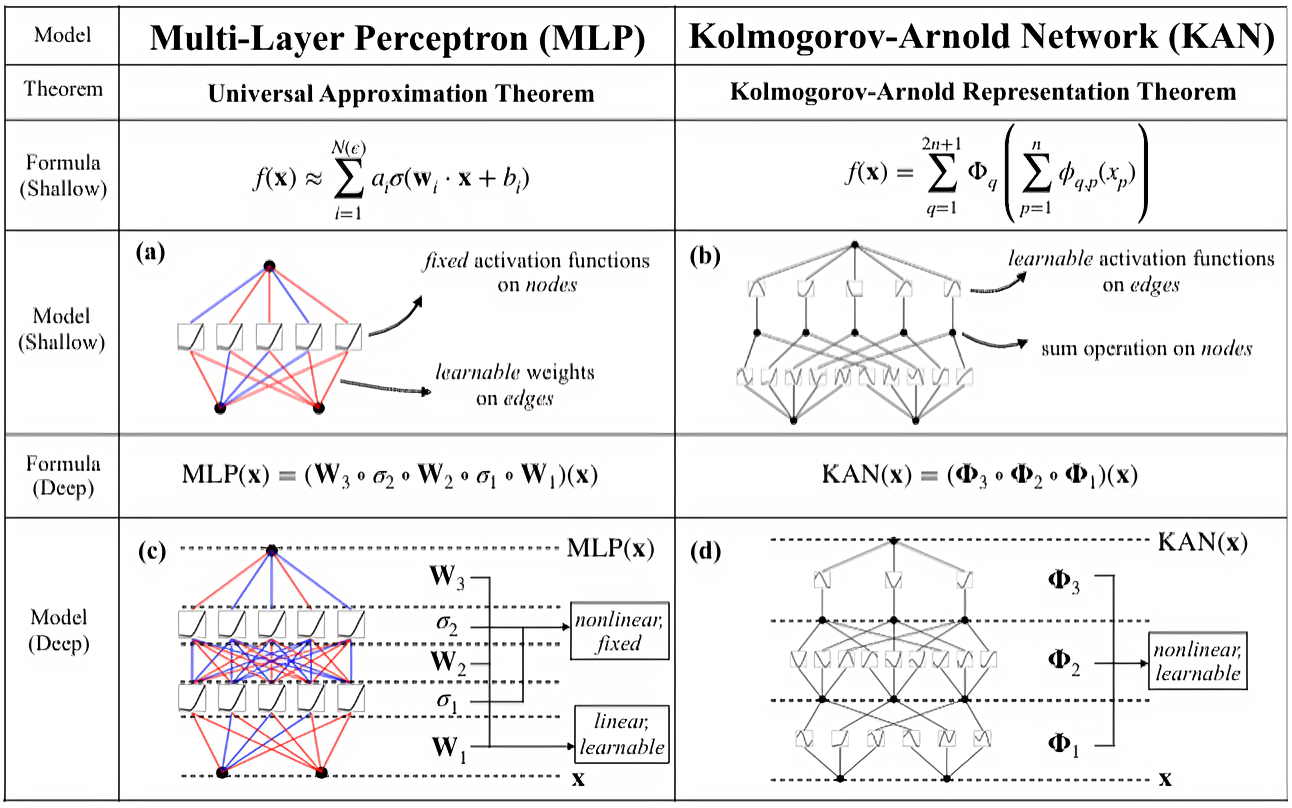}
    \caption{Comparison of  MLP's and KAN's \cite{liu2024kankolmogorovarnoldnetworks}.}
    \label{mlpkan}
\end{figure*}

KANs have emerged as an efficient alternative to MLPs in deep learning due to their ability to approximate nonlinear functions. Figure \ref{mlpkan} compares KAN and MLP performance, emphasizing KAN’s computational efficiency. While MLPs rely on the universal approximation theorem, KANs utilize the Kolmogorov-Arnold representation theorem, which decomposes multivariate functions into simpler univariate ones. By replacing static activation functions with spline-parameterized functions along edges, KANs enhance both precision and interpretability, outperforming MLPs in compositional data tasks.

Traditional function approximation relied on MLPs, but their predefined activation functions limited flexibility, interpretability, and efficiency, particularly for high-dimensional data. KANs address these challenges by leveraging the Kolmogorov-Arnold theorem, enabling adaptable, spline-based activation functions. This architecture excels in compositional data processing, reduces parameter complexity, and mitigates the curse of dimensionality, making KANs a superior choice over MLPs.

\subsection{Kolmogorov Arnold Representation theorem}
\begin{figure}[h]
    \centering
    \includegraphics[width=0.75\linewidth,height=0.35\linewidth]{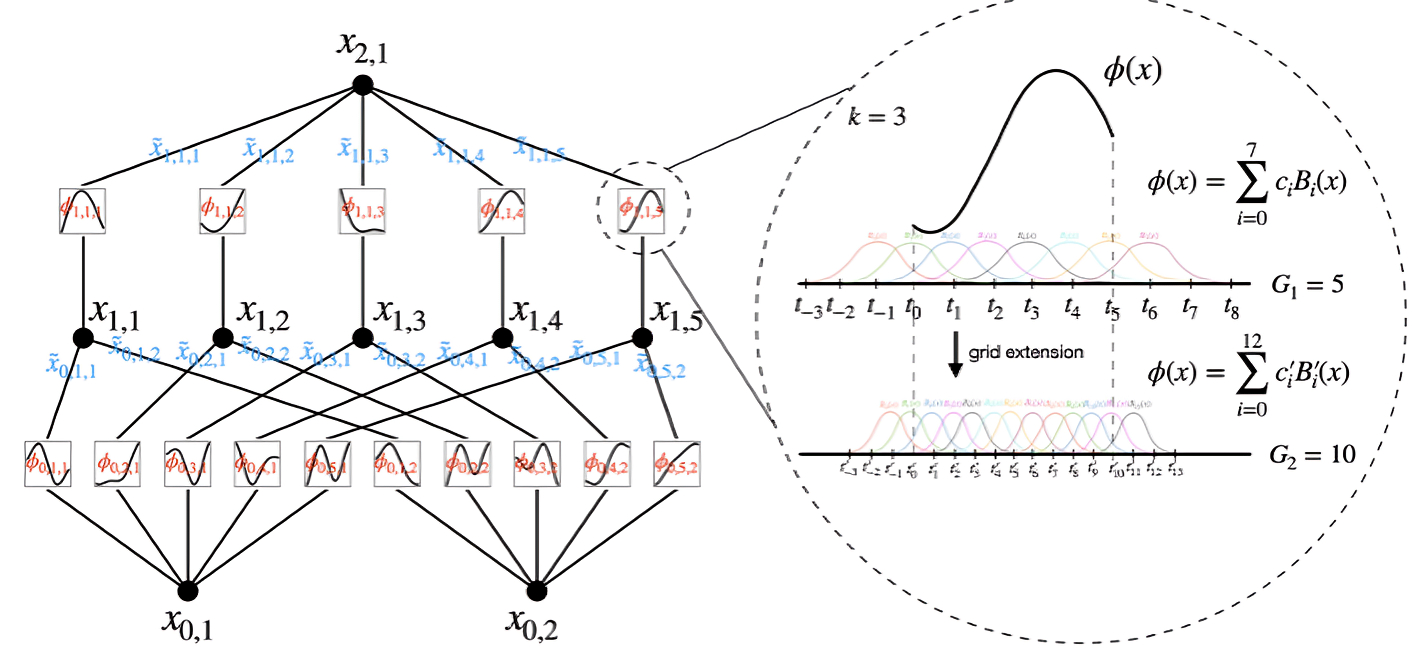}
    \caption{Left: Symbols representing the flow of activations in the network. Right: activation function is defined as
expressed in the form of a B-spline, enabling the transition from coarse to fine grids \cite{liu2024kankolmogorovarnoldnetworks}.}
    \label{fig:kanrepthm}
\end{figure}

KANs relies on the Kolmogorov-Arnold theorem shown in the figure \ref{fig:kanrepthm} which demonstrates the flow of activations in the network, providing insights into the theoretical model, which offers a way to express complex, high-dimensional functions through simpler univariate functions. The theoretical representation, although elegant, was historically considered impractical due to the non-smooth and fractal characteristics of the resulting functions. KANs bring back the formulated theorem by extending its use to deep architectures. KANs offer flexibility and efficiency for modern machine learning by allowing various depths and widths, instead of sticking to the original depth-2, width-2n+1 configuration. KANs combine linear transformations and non-linearities into learnable spline-based transformations, unlike MLPs that separate them into weights and fixed activations. The proposed method allows KANs to improve external structures, like feature learning, and internal accuracy by accurately fitting univariate functions. The equation gives the representation of KAN \ref{eq1:KANrepthm} gives insights of how the features are used in the KAN model. Grasping the theoretical basis of KANs prepares one to assess their actual effectiveness. 
\vspace{-3mm}
\begin{equation}
    f(x) = f(x_1,...,x_n) = \sum_{q=1}^{2n+1}\phi_q(\sum_{p=1}^{n}\phi_{p,q}(x_p))\label{eq1:KANrepthm}
\end{equation}

\subsection{Performance and Empirical Results}
KANs demonstrate exceptional performance across benchmark tasks, surpassing MLPs in convergence speed and accuracy, particularly for compositional functions. Scaling law tests show KANs achieve a test loss reduction proportional to \(N^{-4}\), outperforming MLPs by decomposing complex functions into simpler univariate components, mitigating the curse of dimensionality. 

\begin{figure}[h]
    \centering
    \includegraphics[width=0.65\linewidth,height=0.45\linewidth]{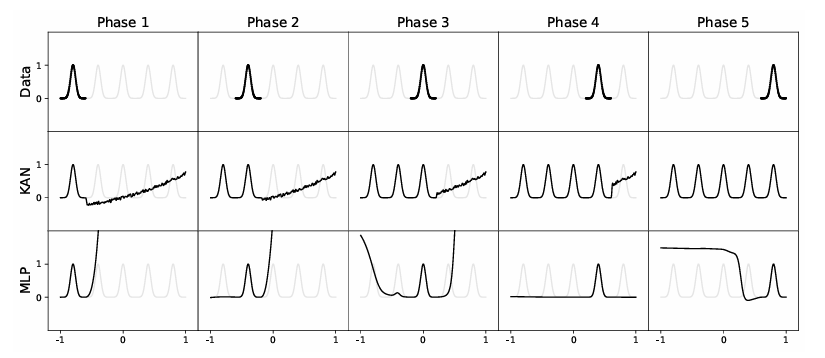}
    \caption{Demonstration of Continual Learning: A KAN model's ability to acquire new tasks while preserving previously learned information
\cite{liu2024kankolmogorovarnoldnetworks}.}
    \label{fig:continuallearning}
\end{figure}

\begin{figure}[h]
    \centering
    \includegraphics[width=0.55\linewidth,height=0.358\linewidth]{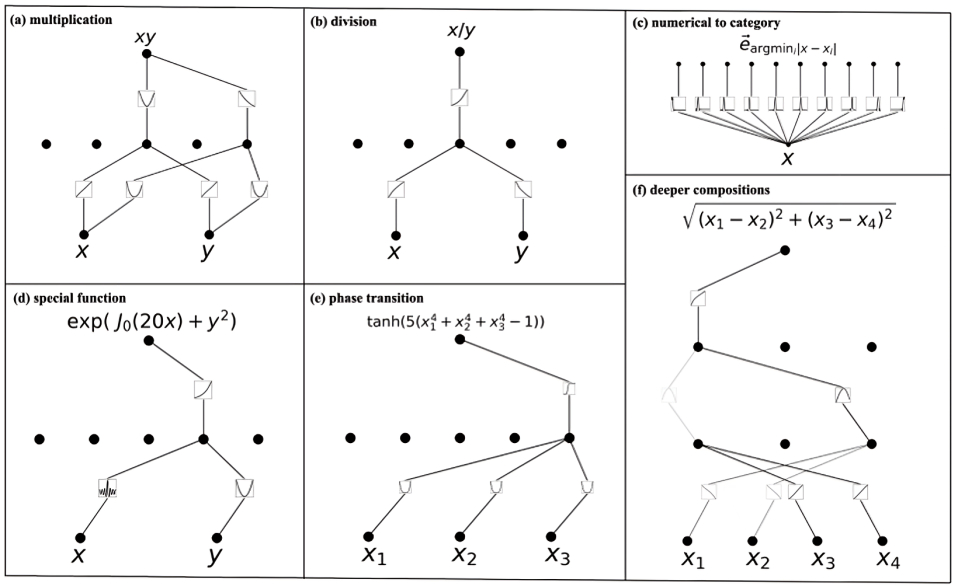}
    \caption{Illustration of Interpretability: A KAN model providing comprehensible insights into decision-making processes
 \cite{liu2024kankolmogorovarnoldnetworks}.}
    \label{fig:interpretable}
\end{figure}

Figure \ref{fig:continuallearning} highlights KANs' ability to retain prior knowledge while learning new tasks, ensuring adaptability in evolving environments. Additionally, Figure \ref{fig:interpretable} illustrates KANs' interpretability, enhancing decision-making processes. Sparsification techniques refine network structures, often yielding symbolic representations like \(sin(x)\) or \(e^{x^2}\), providing direct insights into trained models. This adaptability extends KANs' applicability across diverse learning settings.
 

\begin{figure}[h]
    \centering
    \includegraphics[width=0.75\linewidth,height=0.34\linewidth]{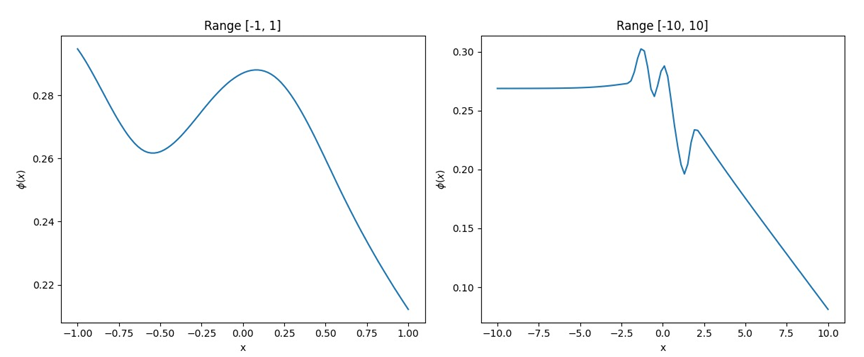}
    \caption{Illustration of Grid extension: Increasing the efficiency of the model, (Bodner et al. \cite{gridextension}).}
    \label{fig:grid}
\end{figure}

In the figure \ref{fig:grid}, Grid extension is again a great feature of KAN. The grid signifies the collection of points utilized for discretization. It is set up according to a specified interval and control points. During the training process, certain input variables for each \(\phi\) could exceed the original range of the grid. To address the issue, the grid is expanded to accommodate inputs that surpass its initial boundaries. The initial KAN describe the extension procedure as an optimization problem where \(G_1\) denotes the initial grid size, \(G_2\) represents enlarged grid size and \(k\) indicates the degree of the B-splines, it is shown in equation \ref{eq2:grid}. During testing, we observed that the output variables of a KAN convolutional layer could exceed the standard grid range of [-1,1]. The proposed test creates a difficulty, particularly with numerous convolutional layers, since the input to a convolutional layer must remain within the operational range of the B-Spline to guarantee that learning is influenced by the splines, rather than solely by the weights adjusting the SiLU activation. To address the proposed approach, during training, the grid is adjusted whenever an input exceeds its range. The proposed modification preserves the original count of control points and spline form while widening the range to include the input. While the process introduces minimal computational overhead, it significantly enhances the model's precision, maintaining KAN's overall efficiency advantage.

\begin{equation}
    c'_j = \arg_{c'_j} \min E_{x \sim p(x)}\left[\sum_{j=0}^{G_2+k-1} c'_j B_j(x') - \sum_{j=0}^{G_1+k-1} c_j B_j(x)\right]
\label{eq2:grid}
\end{equation}
\begin{itemize}
    \item $c_j$ and $c'_j$: Control points of the B-splines representing the original and extended grids, respectively.
    \item $B_j(x)$: B-spline basis functions used to interpolate the input values.
    \item $x$ and $x'$: Input data points within the original and extended ranges, respectively.
    \item $G_1$ and $G_2$: Sizes of the original and extended grids, respectively. $G_2 > G_1$ reflects the expanded grid.
    \item $k$: Degree of the B-splines, controlling the smoothness of the spline curves.
    \item $E_{x \sim p(x)}$: Expectation over the input distribution $p(x)$, representing the likelihood of observing a particular input $x$.
\end{itemize}

\begin{figure}[h]
    \centering
    \includegraphics[width=0.6\linewidth,height=0.5\linewidth]{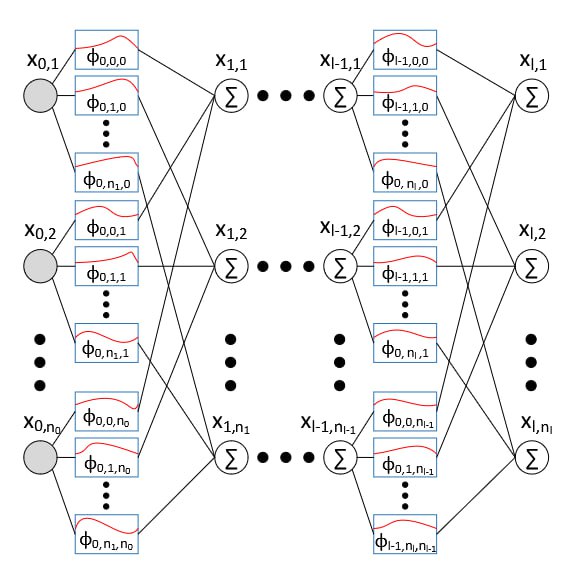}
    \caption{A generalized KAN Architecture, (Tran et al.\cite{pointclass}).}
    \label{fig:pointclass}
\end{figure}

A different approach is to implement batch normalization following every convolutional layer. The method incorporates a limited set of adjustable parameters( \(\mu\) and \(\sigma\)) to standardize the inputs to \(\mu\)=0 and \(\sigma\)=1, making certain that the majority of outputs are within the specified range. Nevertheless, as illustrated in Figure \ref{fig:grid} , when inputs surpass the spline range, the layer behaves similarly to a SiLU activation. If the grid isn't refreshed and many inputs stay outside the range, the KAN essentially operates like an MLP with SiLU activations. Examining the splines acquired throughout various convolutional layers showed no recognizable patterns. Their actions changed greatly based on the layer and position, suggesting that the learning process is influenced by context and does not follow a standardized pattern. Figure \ref{fig:pointclass} offers a generalized framework of KAN, emphasizing its structural benefits compared to traditional neural networks. 

\section{Proposed Methodology}\label{sec:modifications}
Neural networks have demonstrated remarkable capabilities in fields such as image processing, natural language processing, and data classification. However, assessing their effectiveness across different tasks and datasets remains challenging, particularly in resource-constrained environments. Function approximation, time-series forecasting, and classification tasks are fundamental to applications in engineering, finance, and scientific research. While MLPs are widely used, KANs, inspired by Kolmogorov’s theorem, offer an alternative approach for efficiently approximating complex functions.

\begin{figure}[h]
    \centering
    \includegraphics[width=0.35\linewidth,height=0.45\linewidth]{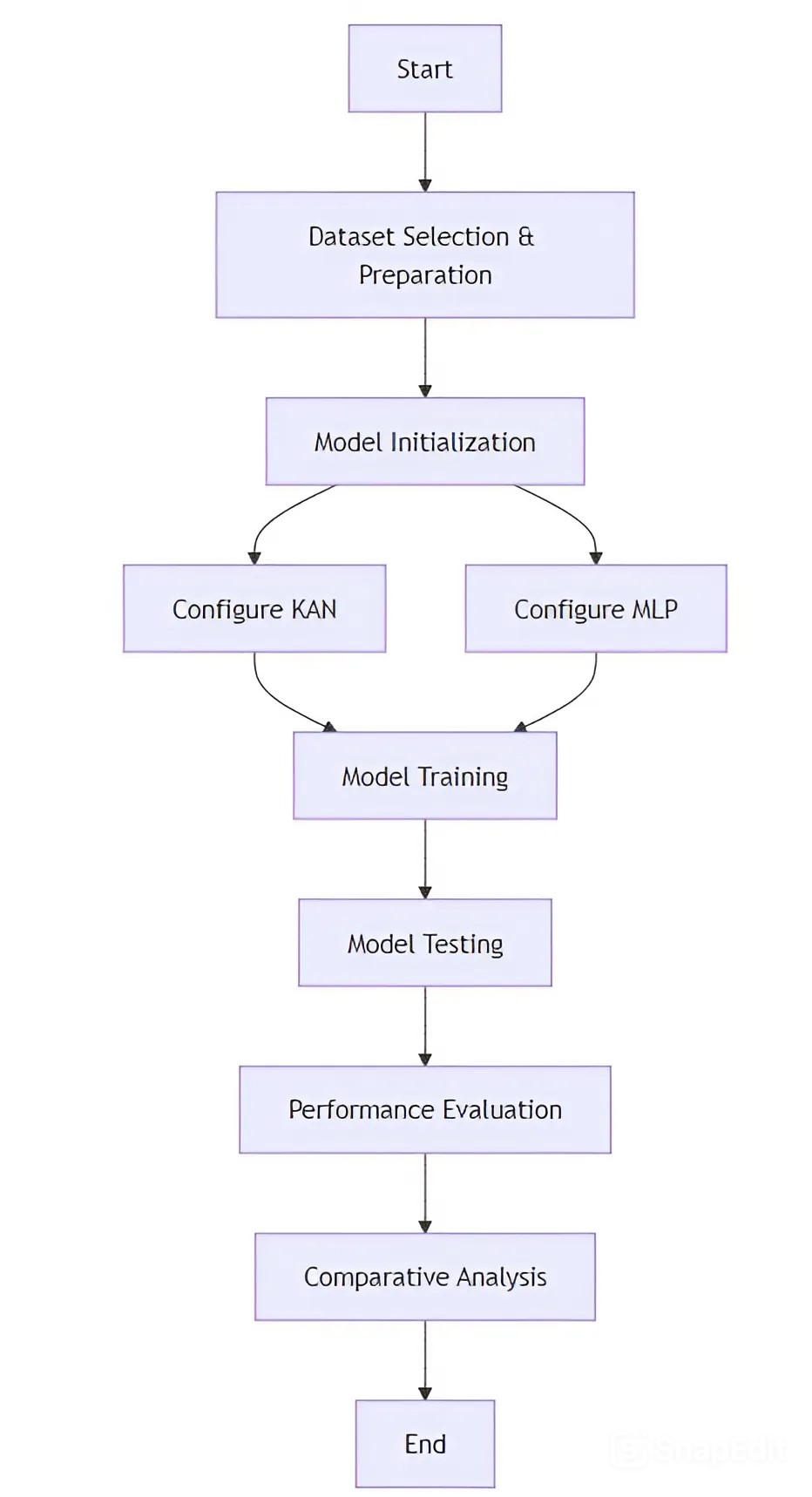}
    \caption{Flowchart of the Proposed Methodology}
    \label{fig:methodology}
\end{figure}

The study examines the relative performance of KANs and MLPs in function approximation, time-series prediction, and classification tasks. Utilizing Kolmogorov’s representation theorem, KANs employ spline-based activation functions along edges rather than nodes, enhancing efficiency and interpretability. Datasets, including square and cube number approximations, daily temperature forecasting, and the Wine dataset for multivariate classification, are analyzed. Architectural differences, optimization strategies, and computational metrics such as MSE and FLOPs are evaluated to assess trade-offs between accuracy and resource consumption. The findings highlight KAN’s improved precision and computational efficiency, emphasizing its potential for resource-limited applications.
Figure \ref{fig:methodology}
is a flowchart of our methodology's steps.
\vspace{-1mm}

\subsection{Architecture of KAN}
KANs have a distinctive architectural structure with activation functions located on edges instead of nodes. The activation function of every edge is defined using a B-spline that can be modified while training. These curves, in their essence, can be adjusted to depict different levels of detail, enabling a hierarchical learning process. Residual connections are integrated into the spline-based activations to enhance training optimization and stability of the network. In contrast to MLPs, KANs achieve a smaller parameter space without depending on linear weights, while still keeping or improving their representational capacity. An important characteristic of KANs is their capability to enlarge spline grids, known as grid extension. Refinement spline activations resolution is done in the presented process without having to retrain the network completely. As the training advances, the grid points specifying the splines become progressively refined, improving the network's capacity to estimate intricate functions. Such a characteristic gives KANs a computational edge, as the network becomes more precise with time, without requiring significant computational resources for retraining. In the equation \ref{eq2: a}, \ref{eq2: b}, \ref{eq2: c}, \( c_i \) is trainable. 
Each activation function is initialized to have \( w_s=1\) and \(\text{spline}(x) \approx 0\).
 \(w_b\) is initialized according to the Xavier initialization.

\begin{equation}
    \phi(x)=w_bb(x) + w_sspline(x)\label{eq2: a}
\end{equation}
\vspace{-0.5cm}
\begin{equation}
    b(x)=silu(x)= x/(1+e^{-x})\label{eq2: b}
\end{equation}
\vspace{-0.5cm}
\begin{equation}
    spline(x)=\sum_ic_iB_i(x)\label{eq2: c}
\end{equation}

\subsection{Dataset Preparation}\label{AA}
The proposed methodology evaluates models using four datasets across time-series and classification tasks. The Square Numbers Dataset, designed to test function approximation, contains 15 rows of numbers and their squares. Similarly, the Numbers Dataset challenges models to predict cube values, also with 15 rows. For time-series analysis, the Daily Minimum Temperatures Dataset includes 3651 rows, each with a date and recorded temperature, requiring models to predict the next day's minimum temperature. The Wine Dataset, widely used in classification, consists of 178 rows with 13 chemical attributes distinguishing wines into three categories. 

Datasets were adapted for both MLP and KAN models. Time-series data was normalized, and categorical labels were one-hot encoded for classification, following Bender et al. \cite{bender2024learningtastemultimodalwine}.

\subsection{Provisioning of MLP Models}\label{AA}
The study first applied MLP models, with fully connected layers tailored to each dataset. Regression tasks used a linear activation in the output layer, while classification employed softmax for probability estimation. In time-series analysis, a sliding window of 3 days predicted the next day's temperature, capturing temporal trends. The Adam optimizer and a learning rate scheduler ensured convergence, with hyperparameters optimized via grid search.

\subsection{Provisioning of KAN Models}\label{AA}
\begin{figure*}[h]
    \centering
    \includegraphics[width=1.0\linewidth,height=0.35\linewidth]{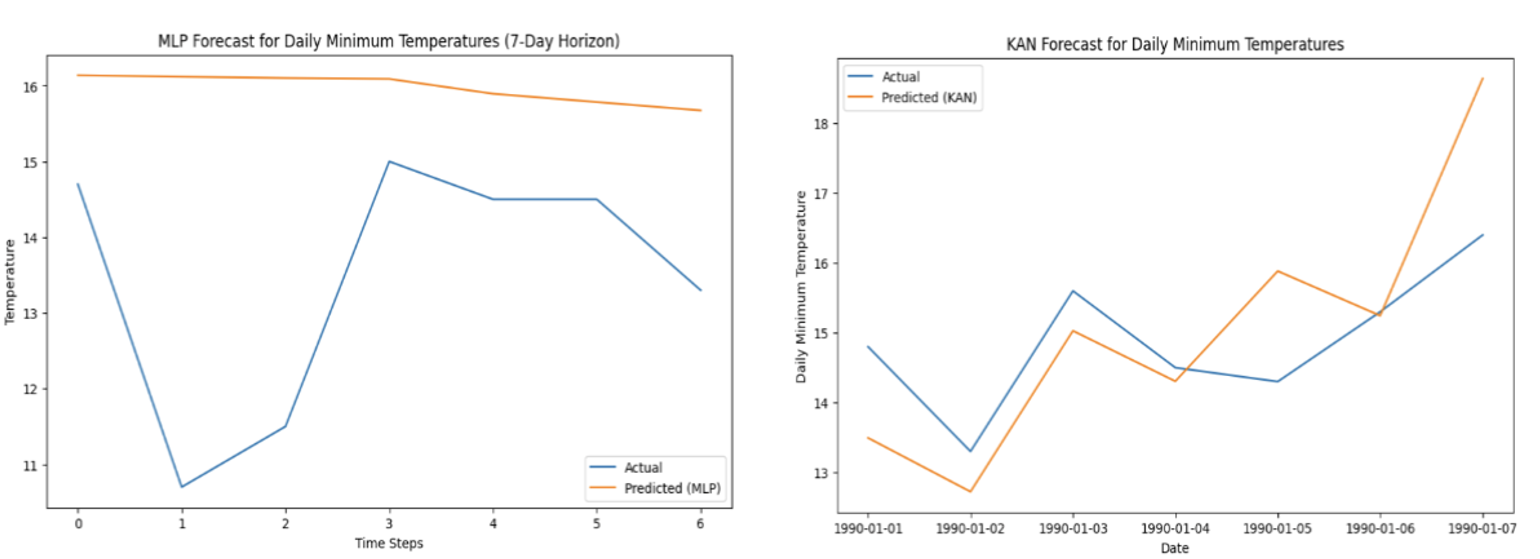}
    \caption{KAN vs MLP (Daily Minimum temperature)}
    \label{fig:kangraph}
\end{figure*}

KAN models were implemented using PyKAN, fastKAN, fasterKAN, and convKAN, leveraging Kolmogorov's theorem to approximate functions through grid-based decomposition. Key parameters, including grid size, layers (k), and spline order, were adjusted per dataset complexity. FastKAN and convKAN were used where computational efficiency was crucial. Grid setups in the daily minimum temperature dataset captured temporal patterns, enabling sequential modeling, as illustrated in Fig \ref{fig:kangraph}.

\subsection{Performance Evaluation and Optimization of Models}\label{AA}
The evaluation of both KANs and MLPs was conducted using essential metrics to guarantee an equitable and thorough comparison. For regression tasks, such as approximating square and cube functions and forecasting daily minimum temperatures, accuracy was evaluated using the MSE. In classification tasks, like sorting wine varieties according to their chemical characteristics, accuracy acted as the main assessment metric. An iterative optimization method was utilized to guarantee fairness. Models demonstrating diminished performance were consistently improved. For MLPs, the modifications included in the layer count, the number of neurons in each layer, and the learning rates. For KANs, the grid dimensions, spline degrees, and layer setups were altered. The repetitive procedure persisted until both models reached similar performance metrics on all datasets. To evaluate computational efficiency, FLOPs were computed for each architecture. For KANs, the De Boor formula was used to determine FLOPs, which quantifies computations involved in spline operations. The formula for FLOPs is given by an interpretation of the de Boor algorithm, as explained by Nava-Yazdani \cite{debooralgorithm}. \(d_{in}\)represents the input dimensions. \(d_{out}\) represents the output dimensions. \(K\) is the dimensionality degree of the activation function.
\begin{equation}
FLOPS=(d_{in}*d_{out})*[9*K*(G+1.5*K]+2*G-2.5*K-1]\label{eq:flops}
\end{equation}
K represents the degree of dimensionality of the activation function. The torchprofile library was used for MLPs to estimate FLOPs, facilitating thorough computations for both single layers and the entire architecture. These assessments guaranteed an equitable comparison of computational efficiency while preserving predictive accuracy. The last step included contrasting the recalibrated FLOPs to determine the optimal design for each dataset. Models with lower FLOPs were considered to be more efficient, since they achieved comparable accuracy while reducing computational costs. The evaluation offered perspectives on the trade-offs between computational cost and forecasting accuracy, emphasizing the advantages and drawbacks of both MLP and KAN models across various datasets. 

\section{Results and Discussion}\label{sec:aabc}
Prior to conducting a thorough performance assessment, it is essential to establish the comparative framework between KANs and MLPs. The findings presented in the document reflect an assessment that includes tasks ranging from basic mathematical function estimation to intricate multivariate classification. The proposed comparative study emphasizes the trade-offs between computational efficiency and accuracy while also pointing out the specific application contexts where each architecture demonstrates its strengths. Such type of investigation offers practical insights into their use in environments with limited resources and high stakes, setting the stage for well-informed design decisions.

\subsection{Perfomance Analysis}
Regarding accuracy and computational efficiency, KAN consistently outperforms MLP for univariate datasets \ref{tab:table1}. For cube value estimation, KAN achieves a lower MSE of 15.2706 versus 2599.5886 for MLP, with a 99\% reduction in computational cost (0.357 kFLOPs vs. 40.000 kFLOPs). Similarly, in square value approximation, KAN attains an MSE of 0.1743 (80.49\% lower than MLP’s 0.8938) while reducing FLOPs by 99.71\%. In time-series tasks, such as predicting daily minimum temperatures, KAN achieves an MSE of 1.4201 (79.87\% lower than MLP’s 7.0565) with a nearly equivalent computational cost (25.284 kFLOPs vs. 24.200 kFLOPs), highlighting its efficiency in sequential data processing. 

For multivariate classification on the Wine dataset, KAN attains 98.43\% accuracy, surpassing MLP’s 96.30\% by 2.21\%, demonstrating its ability to handle complex datasets efficiently. These results emphasize KAN's adaptability in balancing accuracy and computational efficiency. A detailed percentage comparison of FLOPs, MSE, and accuracy across datasets is presented in Table \ref{tab:percentage-diff}.

\captionsetup[table]{skip=10pt}
\begin{table*}[h]
    \centering
    \caption{Comparison of kFLOPS(\(thousands\)) and MSE(\(unit^2\)) for Univariate and Multivariate Datasets}
    \label{tab:table1}
    \renewcommand{\arraystretch}{2.0} 
    \begin{tabular}{|c|c|c|c|c|}
    \hline
    \multicolumn{5}{|c|}{\textbf{Univariate Datasets}} \\
    \hline
    \textbf{Problem} & \multicolumn{2}{|c|}{\textbf{KAN}} & \multicolumn{2}{|c|}{\textbf{MLP}} \\
    \cline{2-5}
                     & \textbf{FLOPS} & \textbf{MSE} & \textbf{FLOPS} & \textbf{MSE} \\
    \hline
    Cube of Numbers & 0.357 & 15.2706 & 40.000 & 2599.5886 \\
    \hline
    Square of Numbers & 0.171 & 0.1743 & 60.000 & 0.8938 \\
    \hline
    Daily Minimum Temperature & 25.284 & 1.4201 & 24.200 & 7.0565 \\
    \hline
    \multicolumn{5}{|c|}{\textbf{Multivariate Datasets}} \\
    \hline
    \textbf{Problem} & \multicolumn{2}{|c|}{\textbf{KAN}} & \multicolumn{2}{|c|}{\textbf{MLP}} \\
    \cline{2-5}
                      & \textbf{FLOPS} & \textbf{MSE} & \textbf{FLOPS} & \textbf{MSE} \\
    \hline
    Point Classification & 4.212 & ACC = 98.43\% & 0.754 & ACC = 96.30\% \\
    \hline
    \end{tabular}
    
\end{table*}

\begin{table}[ht]
\centering
\caption{Percentage variations in FLOPs(thousands), MSE(\(unit^2\)), and classification success between KAN and MLP across various tasks, illustrating performance of models.}
\renewcommand{\arraystretch}{2.0} 
\label{tab:percentage-diff}

\begin{tabular}{|l|l|p{1.5cm}|p{1.5cm}|p{2.5cm}|}
\hline
\textbf{Problem}                & \textbf{Metric}       & \textbf{KAN}    & \textbf{MLP}    & \textbf{Performance Difference} \\ 
\hline

\multirow{2}{*}{Cube of Numbers}     & FLOPs        & 0.357      & 40.000     & 99.11\%               \\ \cline{2-5} 
                                    & MSE             & 15.2706    & 2599.5886  & 99.41\%               \\ \hline

\multirow{2}{*}{Square of Numbers}   & FLOPs       & 0.171      & 60.000     & 99.71\%               \\ \cline{2-5} 
                                    & MSE             & 0.1743     & 0.8938     & 80.49\%               \\ \hline

\multirow{2}{*}{Daily Minimum Temperature}   & FLOPs        & 25.284     & 24.200     & -4.48\%               \\ \cline{2-5} 
                                    & MSE             & 1.4201     & 7.0565     & 79.87\%               \\ \hline

Point Classification                & Accuracy    & 98.43      & 96.30      & 2.21\%                \\ \hline

\end{tabular}
\end{table}

\vspace{-2mm}

\subsection{Accuracy and Computational Efficiency}
The comparative assessment highlights KAN’s superiority in computational efficiency and accuracy. As shown in Table \ref{tab:table1}, KAN consistently outperforms MLP across datasets. For cube estimation, KAN achieves a much lower MSE of 15.2706 compared to MLP’s 2599.5886. Table \ref{tab:percentage-diff} details the percentage differences in FLOPs, MSE, and accuracy, reinforcing KAN’s practical advantages. 

KAN reduces FLOPs by over 99\% in tasks like cube and square approximations, making it ideal for embedded and real-time applications. In time-series tasks, a slight FLOP increase is offset by substantial accuracy gains. For classification, KAN shows marginal accuracy improvements, demonstrating its adaptability and reliability as an alternative to MLP.

\vspace{-2mm}

\subsection{Relevance and Impact Analysis}
KAN's unique architecture, featuring spline-based activation and grid-based designs, enables it to outperform MLP in accuracy and efficiency. Achieving lower MSE with fewer resources makes it ideal for resource-limited environments. Its effectiveness in time-series forecasting and multivariate classification underscores its adaptability. These findings pave the way for further exploration of KAN’s potential in high-dimensional and large-scale applications.

\subsection{Comparison with Previous Approaches}
Compared to MLP, a widely used choice for function approximation and classification, KAN offers clear advantages. While MLP’s fully connected layers capture data patterns, they increase computational complexity. KAN, leveraging Kolmogorov’s theorem, simplifies function approximation, achieving similar or better accuracy with reduced computational demands. This study builds on prior research, positioning KAN as a viable MLP alternative in deep learning. Its efficiency in nonlinear function approximation highlights its potential for multivariate classification and large-scale data modeling.

\vspace{-2mm}
\section{Conclusion and Future Work}\vspace{-3mm}
The study compares KAN and MLP in function approximation, classification, and time-series forecasting. KAN outperforms MLP in efficiency and accuracy, achieving lower MSE with fewer FLOPS, making it ideal for resource-limited environments. Based on Kolmogorov’s theorem, KAN provides a superior solution for non-linear function approximation and sequential data tasks.

KAN effectively handles complex, non-linear data, making it suitable for real-time applications like financial forecasting, robotics, and biomedical signal analysis. Its low computational cost benefits embedded systems and edge AI. Further efficiency gains can be achieved through pruning, quantization, and FPGA acceleration.

%
%
\bibliographystyle{plain}
\bibliography{ref}

\end{document}